\documentclass[letterpaper, 10 pt, conference]{ieeeconf}  

\usepackage[space]{cite}
\usepackage{amsmath,amsfonts}
\usepackage{algorithm}
\usepackage{array}
\usepackage{textcomp}
\usepackage{stfloats}
\usepackage{url}
\usepackage{verbatim}
\usepackage{graphicx}
\usepackage{algpseudocode}
\usepackage{booktabs}
\usepackage{multirow}
\usepackage[switch,pagewise]{lineno}
\usepackage{orcidlink}
\usepackage[utf8]{inputenc}
\usepackage[caption=false]{subfig}
\usepackage{soul,xcolor}
\usepackage{tablefootnote} 
\graphicspath{{./img/}}

\IEEEoverridecommandlockouts                              

\title{\LARGE \bf
    PB-NBV: Efficient Projection-Based Next-Best-View Planning Framework for Reconstruction of Unknown Objects
}

\author{Zhizhou Jia, Yuetao Li, Qun Hao, and Shaohui Zhang 
\thanks{
This paper has been accepted for publication in IEEE Robotics and Automation Letters (RA-L), 2025.
DOI: 10.1109/LRA.2025.3573631.}
\thanks{All authors are with the School of Optics and Photonics, Beijing Institute of Technology, Beijing 100081, China. (Corresponding author: Shaohui Zhang.)
        {\tt\small \{jiazhizhou, liyuetaochn, qhao, zhangshaohui\}@bit.edu.cn}}%
}

\begin{document}
\maketitle
\thispagestyle{empty}
\pagestyle{empty}

\begin{abstract}
Completely capturing the three-dimensional (3D) data of an object is essential in industrial and robotic applications.
The task of next-best-view (NBV) planning is to calculate the next optimal viewpoint based on the current data, gradually achieving a complete 3D reconstruction of the object.
However, many existing NBV planning algorithms incur heavy computational costs due to the extensive use of ray-casting.
To address this, we propose a projection-based NBV planning framework. 
Specifically, this framework refits different types of voxel clusters into ellipsoids based on the voxel structure.
Then, the next optimal viewpoint is selected from the candidate views using a projection-based viewpoint quality evaluation function in conjunction with a global partitioning strategy.
This process replaces extensive ray-casting, significantly improving the computational efficiency.
Comparison experiments in the simulation environment show that our framework achieves the highest point cloud coverage with low computational time compared to other frameworks.
The real-world experiments also confirm the efficiency and feasibility of the framework.
Our method will be made open source to benefit the community \footnote{ \href{https://github.com/dspangpang/pb\_nbv}{To be released at:https://github.com/dspangpang/pb\_nbv}}.
\end{abstract}

\section{Introduction}
\label{sec:intro}

{A}{cquiring} the complete and high-quality 3D geometric structure of an object is crucial for reverse engineering and quality inspection in industrial applications, as well as for autonomous exploration and interaction in robotics.
In practical scanning, achieving a complete scan can be challenging due to occlusions or limitations in the acquisition device's field of view (FOV).
In response, Connolly \cite{connolly1985determination} first introduced a strategy that leverages the observed data to identify the next optimal scanning viewpoint for achieving a comprehensive object scan, known as the NBV planning problem.
As research on the NBV planning problem advanced, it was gradually modeled into three core components: object representation, candidate viewpoint proposal, and optimal viewpoint selection.

In many existing algorithms, once the object is represented by a voxel structure, the ray-casting algorithm is typically used to assess the visibility of each candidate viewpoint to the object \cite{pan2021global,potthast2014probabilistic,batinovic2021multi,batinovic2022shadowcasting}.
This algorithm involves emitting multiple rays from each candidate viewpoint and examining the occlusion of each ray with the target structure.
However, this process involves numerous queries, and as the object's volume and precision requirements increase, the use of the ray-casting algorithm imposes a significant computational burden \cite{mendoza2020supervised, pan2022scvp, border2024surface}, limiting the deployment of the NBV planning algorithm on lightweight edge computing robot platforms.
\begin{figure}[t]
    \centering
    \includegraphics[width=3.4in]{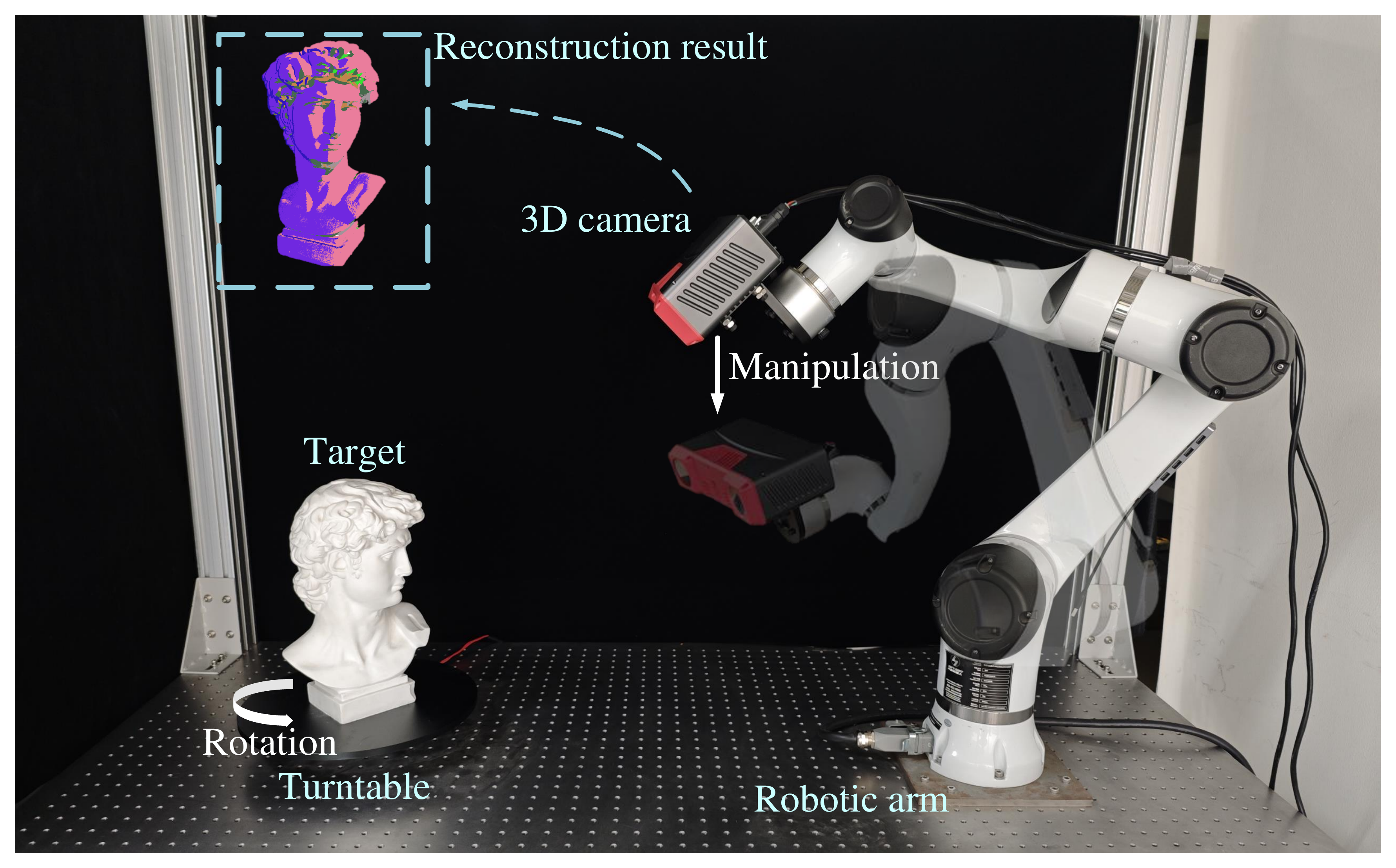 }
    \caption{Overview of the NBV planning experiment platform. An object to be measured is placed on a turntable, and a 3D camera is equipped at the end of the robotic arm for data acquisition. Our NBV planning framework accomplishes a complete reconstruction of the object by controlling the robotic arm and the turntable.
    }
    \label{Experimental_platform}
\end{figure} 

To address this issue, we propose a novel projection-based NBV planning framework.
The framework is based on voxel structures, where the voxels are further fitted into ellipsoids. 
When selecting the next optimal viewpoint, an ellipsoid-based independent projection strategy for evaluating observation quality replaces ray-casting, significantly reducing the computational burden caused by an increase in object volume or higher precision requirements for the representation structure.
The framework also introduces a global partitioning strategy to ensure the success of point cloud registration between frames and reduce backtracking caused by greedy viewpoint selection.

To evaluate the performance of our framework, we set up a simulation environment for ablation studies and comparisons with existing mainstream NBV frameworks. Additionally, the real-world experiments also confirm the efficiency and feasibility of the framework. The experiments were conducted in a real-world setup similar to SEE\cite{border2024surface}, as shown in Fig. \ref{Experimental_platform}.


In summary, the main contributions of this paper are the following three:

1) We propose an ellipsoid-fitting-based representation structure for unknown objects, enabling dimensionality reduction of voxel structures while retaining essential information.

2) We propose an ellipsoid-based projection algorithm to replace the ray-casting in selecting the next optimal viewpoint, significantly reducing the computational burden.
  
3) We introduce a global partitioning strategy to ensure point cloud registration success and reduce the impact of backtracking from greedy selection on efficiency.

\section{Related Works}
\label{sec:related}

Currently, the NBV planning problem is typically modeled into three core components: object representation, candidate viewpoint proposal, and optimal viewpoint selection. Research has proposed various solutions and strategies by optimizing and innovating these components.

\subsection{Object Representation}
Object representation is a fundamental aspect of the NBV planning problem; different representation methods directly affect the accuracy and efficiency of the overall NBV planning framework.
The object representation, ranging from simple to complex, typically include point clouds\cite{border2024surface, zeng2020pc,dhami2023pred}, voxels\cite{connolly1985determination,pan2021global,potthast2014probabilistic,vasquez2014view,pan2022scvp,mendoza2020supervised,song2020online,zhou2021fuel,batinovic2021multi,vasquez2018tree,deng2020frontier,batinovic2022shadowcasting}, and surface\cite{roberts2017submodular,peng2019adaptive,wu2014quality}structures.
With the advancement of computer graphics, emerging structures such as Implicit Neural Representations (INR)\cite{ran2023neurar,lee2022uncertainty, jin2023neu} are also being utilized to describe NBV planning problems.
Voxel structures are commonly used for object representation due to their simplicity, intuitiveness, and controllable precision.

\subsection{Candidate Viewpoint Proposal}
Finding an optimal 6-DOF viewpoint directly through optimization in space is very challenging, so researchers usually sample a number of candidate viewpoints and then choose the best one.
Currently, common candidate viewpoint sampling strategies can generally be categorized into four types: random sampling\cite{pan2021global,potthast2014probabilistic}, predefined set\cite{connolly1985determination,mendoza2020supervised,pan2022scvp}, path-based sampling\cite{zhou2021fuel,vasquez2018tree,deng2020frontier,batinovic2022shadowcasting}, and scene-information-based sampling\cite{border2024surface,batinovic2021multi}.
The candidate viewpoint sampling strategy should be appropriately adjusted according to the current holistic data acquisition system.
An appropriate strategy will enhance the convergence efficiency of the NBV planning framework.

\subsection{Optimal Viewpoint Selection}
It's hard to tell which areas have more potential information just from the object's structure(which only includes position data).
To measure this information, researchers usually classify each structural unit by the sensor's observation pose; different units indicate varying levels of potential information in their areas.
The strategy for selecting candidate viewpoints aims to maximize this potential information.

Optimal viewpoint selection strategies for different NBV problems and scenarios can be divided into those focusing on completeness and those focusing on efficiency.
\subsubsection{Completeness-focused strategies}
Focusing on voxel structures, the optimal viewpoint selection strategies that emphasize completeness are primarily divided into position-based classification\cite{song2020online, zhou2021fuel, batinovic2021multi, batinovic2022shadowcasting,deng2020frontier}, occupancy probability-based classification\cite{potthast2014probabilistic}, and a combination of both\cite{pan2021global}.

Position-based classification focuses on edge information and classifies frontier units representing edge locations from all representation units. The optimal viewpoint should capture the maximum number of Frontier units. 
The occupancy probability method determines the probability of each representation unit being a solid object, using information entropy theory to judge which viewpoint will bring the greatest information gain. 
The combined strategy first classifies the representation units, then assigns different occupancy probabilities to different units, and finally selects the optimal viewpoint through information gain.
Also, some strategies incorporate practical environmental constraints and use utility functions to adjust greedy indicators, preventing robots from being unable to reach certain viewpoints\cite{vasquez2014view}.

\subsubsection{Efficiency-focused strategies}
In the process of selecting the optimal viewpoint, the potential information of each candidate viewpoint must be calculated iteratively. 
The commonly used ray-casting algorithm judges the visibility of units by emitting multiple rays, but it poses a significant computational burden when dealing with complex structures. 
Efficiency strategies aim to address these issues.

The first category of strategies focuses on simplifying the representation structure by directly analyzing potential information at the point cloud level\cite{border2024surface,zeng2020pc,dhami2023pred}, thereby reducing the time spent on structure construction and visibility queries.

The second category of strategies enhances the efficiency of ray-casting.
Both \cite{vasquez2014view} and \cite{batinovic2022shadowcasting} propose strategies to improve the efficiency of ray-casting itself. \cite{monica2018surfel} uses the GPU to accelerate ray-casting.
Some deep learning strategies can avoid using ray-tracing algorithms and provide the best viewpoint at a fast speed.
\cite{mendoza2020supervised} and \cite{zeng2020pc} use the network to predict the information gain for selecting the next perspective. \cite{pan2022scvp} can infer all subsequent observation viewpoints by inputting only one frame of observation.
However, generalizing such algorithms to other unknown objects that differ from their training set is challenging \cite{border2024surface}.

Inspired by previous research and the 3D Gaussian Splatting method\cite{kerbl20233d}, we propose an NBV planning framework based on voxel structures. 
The framework consolidates classified frontier and occupied voxels into an ellipsoid and evaluates the potential observation quality of viewpoints through projection.
It completely replaces ray-casting in the process of candidate viewpoint evaluation, significantly improving computational efficiency. 
The next section will elaborate on the specifics of this framework in detail.

\section{Methodology}
\label{sec:method}

\begin{figure*}[t]
    \centering
    \includegraphics[width=7in]{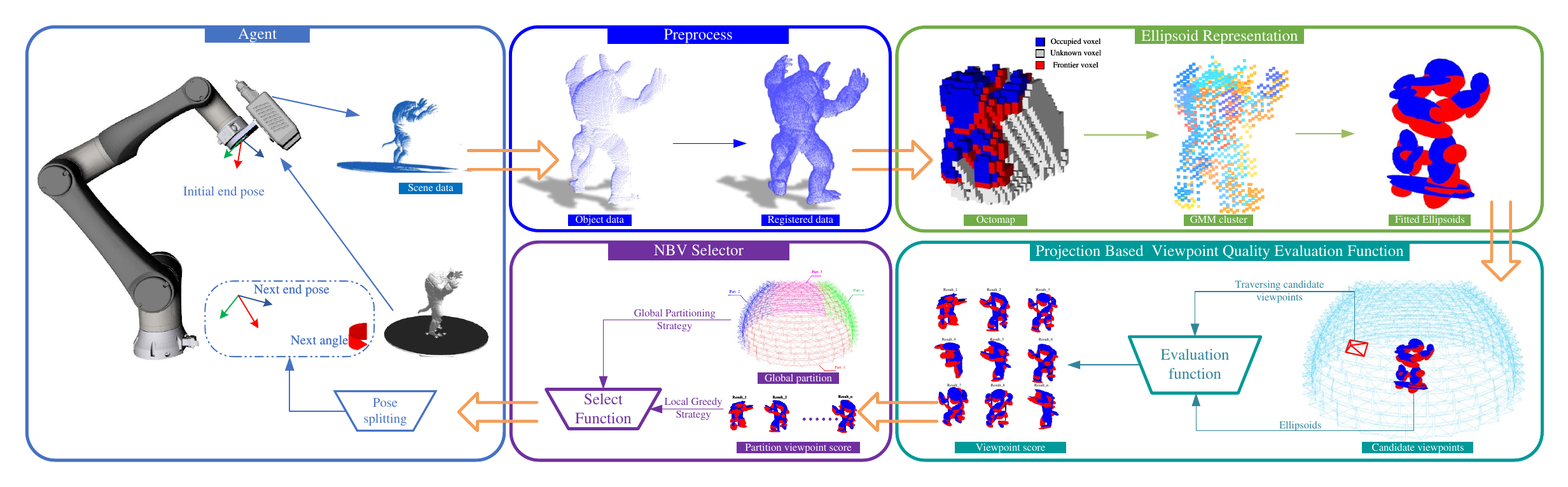}
    \caption{Overview of NBV planning framework. The orange arrows describe the running steps of the NBV iteration process. 
    After capturing the point cloud, the framework performs preprocessing and pose registration. The registered point cloud is then input into the ellipsoid representation module, where it is converted into a voxel structure and fitted into an ellipsoid. The projection evaluation function is used to assess all candidate viewpoints, and a global partitioning strategy selects the optimal viewpoint for the next frame. Finally, the robotic arm moves to the selected viewpoint and begins the next iteration.}
    \label{System_Overview}
    \vspace*{-1\baselineskip} 
\end{figure*}

The workflow of our projection-based NBV planning framework is shown in Fig. \ref{System_Overview}. 
The robot arm first moves to a preset initial observation pose and then starts the iterative process of NBV planning.

\subsection{Proposal of Candidate Viewpoints}
As shown in Fig. \ref{Experimental_platform} of the experimental setup, we assume that the bounding box radius of the object to be reconstructed will not exceed the distance from the robotic arm base to the turntable. Considering the depth of field limitations of the 3D camera, we propose a strategy to dynamically adjust the candidate view sampling area to ensure complete reconstruction of the object within the optimal shooting distance of the 3D camera.

\begin{figure}[t]
    \centering
    \includegraphics[width=3.4in]{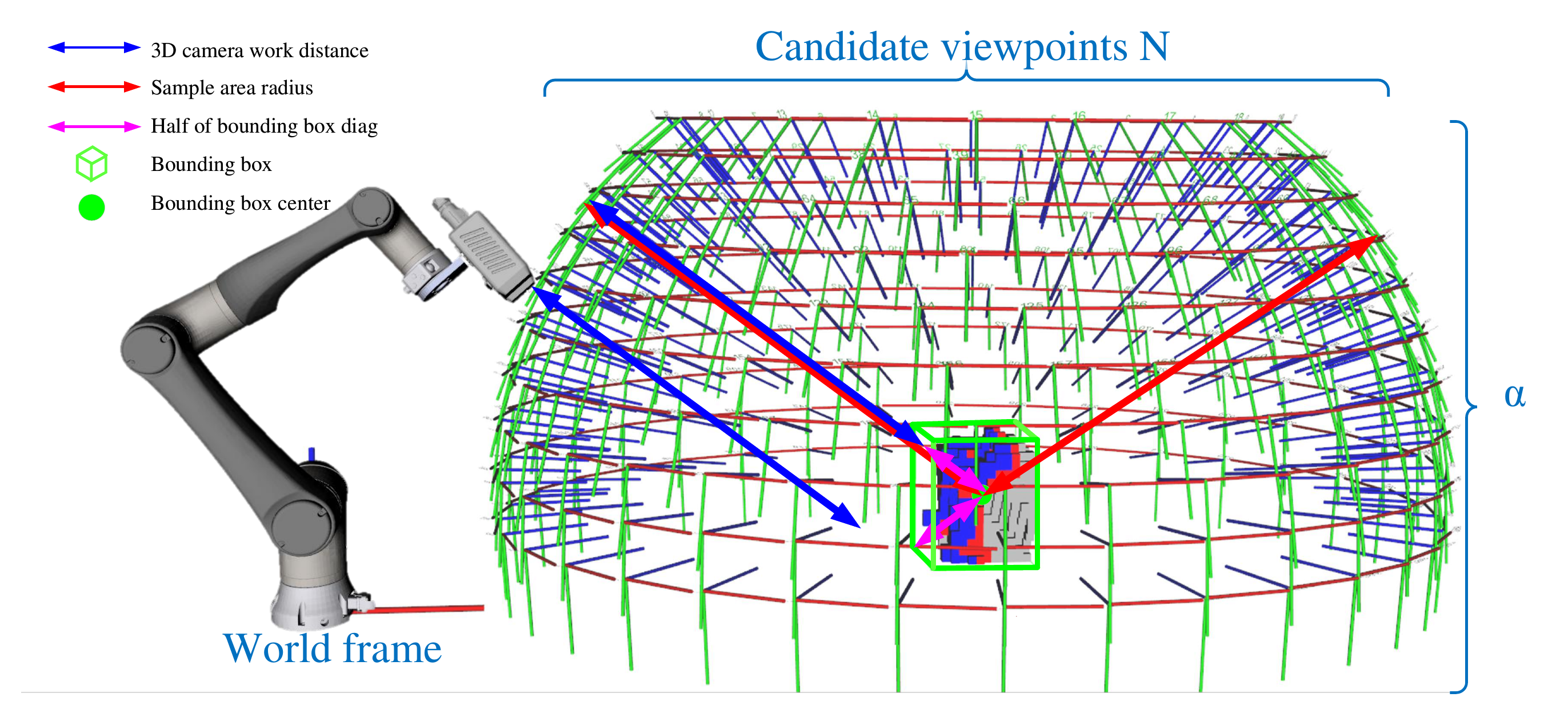}
    \caption{The components of the radius of the candidate viewpoint sampling region and the results of candidate viewpoint sampling within this region.}
    \label{Proposal_of_Candidate_Viewpoints}
    \vspace*{-1\baselineskip} 
\end{figure}

Considering the reachability of the robot arm, we set the sampling area of the candidate viewpoint to the partial hemisphere area above the turntable.
The algorithm uses the base coordinates of the robot as the world coordinate system, as shown in Fig. \ref{Proposal_of_Candidate_Viewpoints}. 
The radius of this hemisphere $R$ is the sum of the 3D camera's working distance  $d_c$ and half the diagonal length of the unknown object's bounding box $d_b$. 
Inspired by unknown-target scanning strategies from \cite{daudelin2017adaptable, pan2024integrating}, the hemisphere dynamically resizes with the object's bounding box to fit the 3D camera's depth of field.
We selected \(\alpha\) equidistant parallels from the candidate area and distributed \(N\) candidate viewpoints proportionally along each parallel based on their lengths.
All viewpoints are directed toward the center of the bounding box.

\subsection{Voxel Structure Construction}
\begin{figure}[t]
    \centering
    \subfloat[]{\includegraphics[width=1.6in]{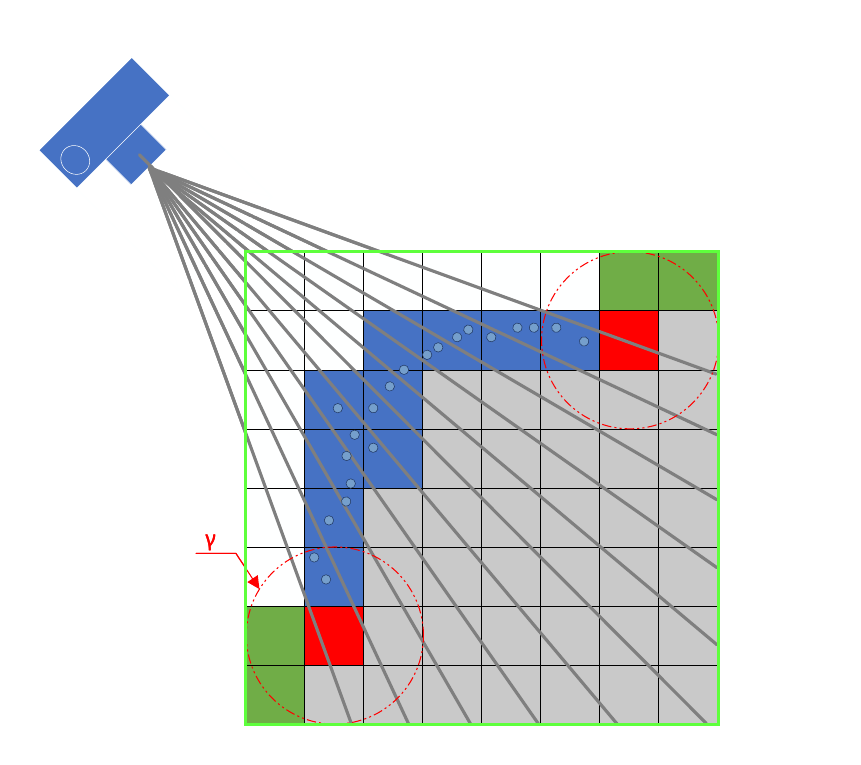}
    \label{Voxel_struct_a}
    }
    \subfloat[]{\includegraphics[width=1.6in]{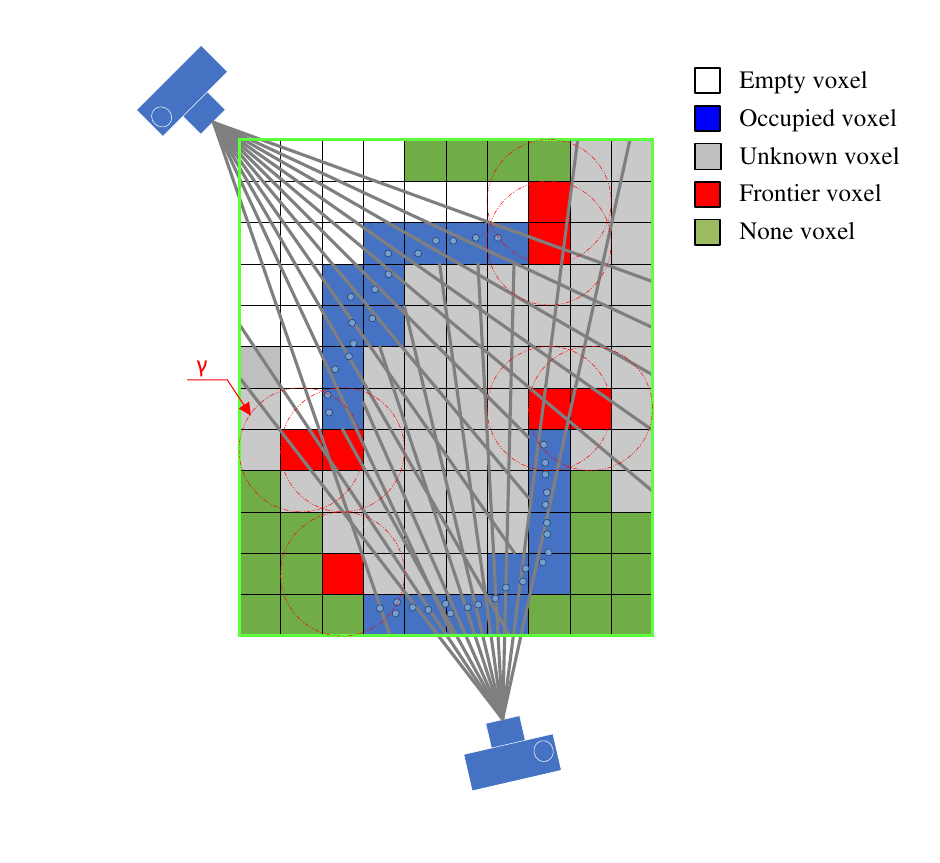}
    \label{Voxel_struct_b}
    }
    \caption{Use a 2D grid to describe the voxel classification. (a) Classification results of a single frame input. (b) Classification results of multiple frames input. The green box represents the object's bounding box.}
    \label{Voxel_struct}
    \vspace*{-1\baselineskip} 
\end{figure}

After the 3D camera captures a point cloud frame, the framework first performs preprocessing on the data.
The preprocessed point cloud is then registered into the point cloud set \(P_f\) and subjected to voxel construction.
We categorize voxels within the unknown object's bounding box into five types: Empty \(V_e\), Occupied \(V_o\), Unknown \(V_u\), Frontier \(V_f\), and None \(V_n\), as shown in Fig. \ref{Voxel_struct}.

All newly generated voxels are initialized to state $V_n$. Upon completion of sensor observation, multiple ray bundles are cast toward the target region through ray-travel\cite{amanatides1987fast}, with voxel states updated according to the following rules: 
1) If point cloud data $P_f$ resides within a $V_n$ voxel, the voxel is updated to $V_o$, indicating an occupied region of the target surface; 
2) When a $V_o$ voxel exists along the ray path, all $V_n$ voxels penetrated by the ray behind this $V_o$ voxel are labeled as $V_u$, representing occlusion-induced uncertainty; 
3) All $V_n$ voxels along the ray path prior to encountering any $V_o$ voxel are updated to $V_e$, denoting observed free space.
4) The update of $V_f$ is carried out after the updates of $V_u$ and $V_o$. Any $v_i\in V_u$ is identified as a Frontier voxel if both empty and occupied voxels are present in its spatial neighborhood.
\begin{equation}
    V_f = \{v_u^{i} \mid \exists  v_e \in \mathcal{N}(u^{i}) \And v_o \in \mathcal{N}(u^{i})  \}
\end{equation}
where $v_u^{i} \in V_u$ and $\mathcal{N}(u^{i}) $represents the spatial neighborhood of $v_u^{i}$.

The bounding box $\mathbf{B}$ is initialized to the bounding box of $V_o$ in the first frame.
Then, $\mathbf{B}$ is expanded along the direction of the current viewpoint, doubling its diagonal length. 
In subsequent updates, a sphere $s_i$ with a radius of $\gamma$ is generated around each voxel $v_i \in V_f$, and the size of $\mathbf{B}$ is expanded to cover $V_o$, $V_u$, and $s_i$. 
The result is shown in Fig. \ref{Voxel_struct}\subref{Voxel_struct_b}.

\subsection{Ellipsoid Representation}

\begin{figure}[t]
    \centering
    \includegraphics[width=3.4in]{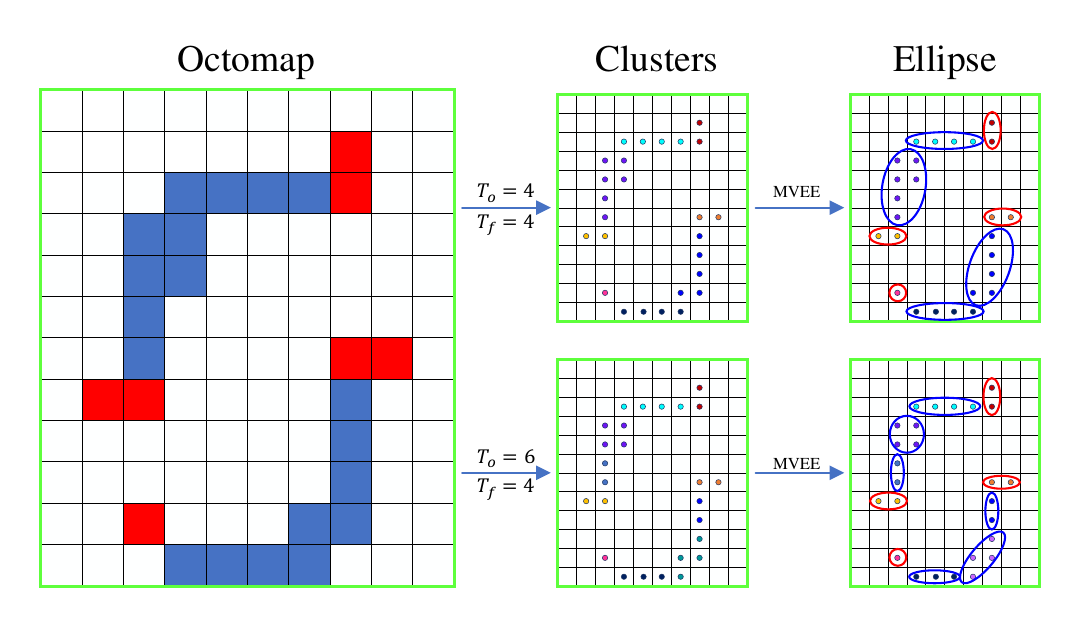}
    \caption{Results of different GMM clustering numbers. \( T_o \) represents the number of clusters of occupied voxels, and \( T_f \) represents the number of clusters of frontier voxels.}
    \label{Ellipsoid_representation}
    \vspace*{-1\baselineskip} 
\end{figure}

In 3D games, an entity's collision structure can be represented by an ellipsoid\cite{fauerby2003improved}, and more complex structures can be achieved by using multiple ellipsoids for different parts of an entity.
Inspired by this, we also use multiple ellipsoids to describe an object structure. 
Since the object is unknown and its structural characteristics are unclear, it is necessary to roughly divide the object into \( t \) clusters and fit each cluster into an ellipsoid to represent the object structure. 
Fig. \ref{Ellipsoid_representation} illustrates this process using a two-dimensional diagram.

Inspired by 3D Gaussian Splatting\cite{kerbl20233d}, we know that the equiprobability density surface of a 3D Gaussian distribution is an ellipsoid. Suppose a set of points in space follows a 3D Gaussian distribution. In that case, its covariance matrix can indicate the dispersion of these points along each axis, making it easy to fit these points into an ellipsoid. Hence, we will employ a Gaussian Mixture Model (GMM) for voxel clustering.

Occupied voxels and frontier voxels will directly affect the viewpoint quality evaluation results.
Therefore, after each update of $V_o$ and $V_f$, we will initialize $T$ with 3D Gaussian distributions for each type of voxel ($V_o$ and $V_f$).
We utilize the expectation-maximization (EM) algorithm to train a Gaussian distribution, which allows us to obtain the clustering results $C_o$ and $C_f$ after categorizing these two types of voxels.

The selection of the number of Gaussian distributions $T$ will directly affect the number of fitted ellipsoids in the subsequent algorithm. 
Too few ellipsoids will make it difficult to accurately reflect the local structure of the object, while too many can impact the efficiency. 
To allow the algorithm to select the number of Gaussian distributions adaptively, we use the Bayesian Information Criterion (BIC)\cite{schwarz1978estimating} to evaluate the model parameters.
\begin{equation}
    \text{BIC} = k \ln (n) - 2 \ln *(L)
\end{equation}
$k$ is the number of model parameters, $n$ is the number of samples, and $L$ is the likelihood function.

The core concept of BIC is to find the best trade-off between the number of model parameters and the likelihood function value. 
To balance efficiency and accuracy, we set the value range of $T$ to [0, $T_{max}$], and $T_{max}$ is the manually set maximum ellipsoid number to be fitted. 
Among the two voxels $V_o$, $V_f$, select $T_o$, $T_f$ with the smallest corresponding BIC value as the number of Gaussian distributions $T_{*}$ in the current iteration, that is:
\begin{equation}
    T_{*} = \arg \min_{T_{max}}BIC(GMM(V,T))
\end{equation}

After obtaining $C_o$ and $C_f$, we use the method\cite{gartner1997smallest} in the CGAL library to calculate the minimum-volume enclosing ellipsoids (MVEE) model.
Then, we can obtain $E_o$ and $E_f$, representing the fitting results of the occupied voxels and Frontier voxels.
At this point, we can transform the representation of the object from a voxel structure to an ellipsoid structure.
Each ellipsoid is described by its equation.

\subsection{Projection-based Viewpoint Quality Evaluation Function}

The goal of the viewpoint quality evaluation function proposed in this paper is to enable the viewpoint to observe the most frontier information. 
In this process, we must consider the occlusion problem between structures when observing the object.
Although ray-casting can accurately give the occlusion relationship of each voxel, the extensive use of ray-casting will take up a lot of computing resources.
To address this, we use a new strategy that calculates the observable weight by using the order of the center positions of each ellipsoid under the current viewpoint. 
Then, the final viewpoint evaluation result is obtained by accumulating the results of individually weighted projections of each ellipsoid.
The details of evaluating a single viewpoint can be divided into three parts: ellipsoid observation weight calculation, ellipsoid weighted projection calculation, and viewpoint quality result calculation.

\begin{figure*}[!t]
    \centering
    \subfloat[]{\includegraphics[width=1.1in]{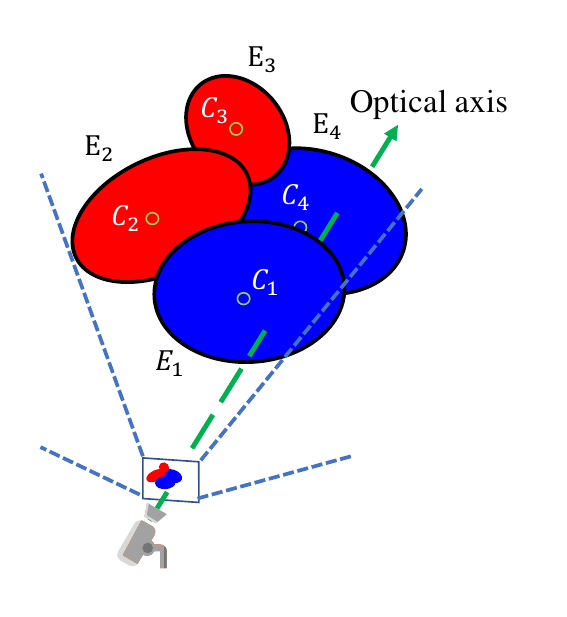}
    \label{Viewpoint_quality_evaluation_function_a}
    }
    \subfloat[]{\includegraphics[width=2.2in]{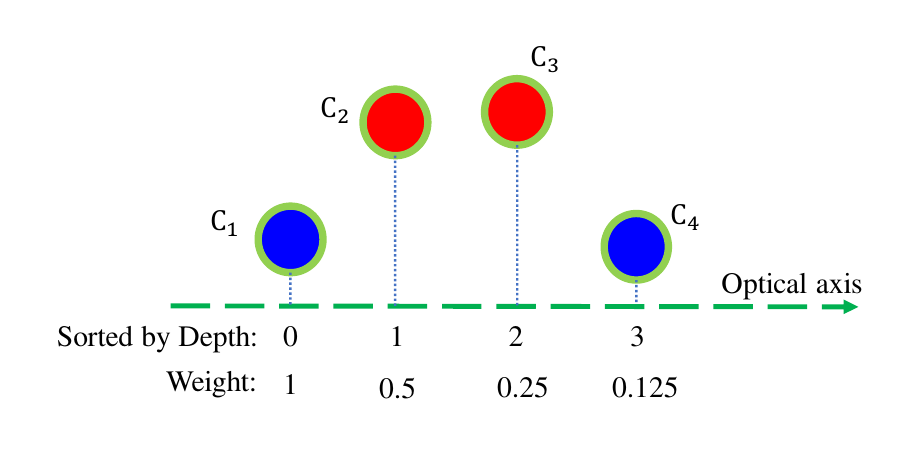}
    \label{Viewpoint_quality_evaluation_function_b}
    }
    \subfloat[]{\includegraphics[width=2.2in]{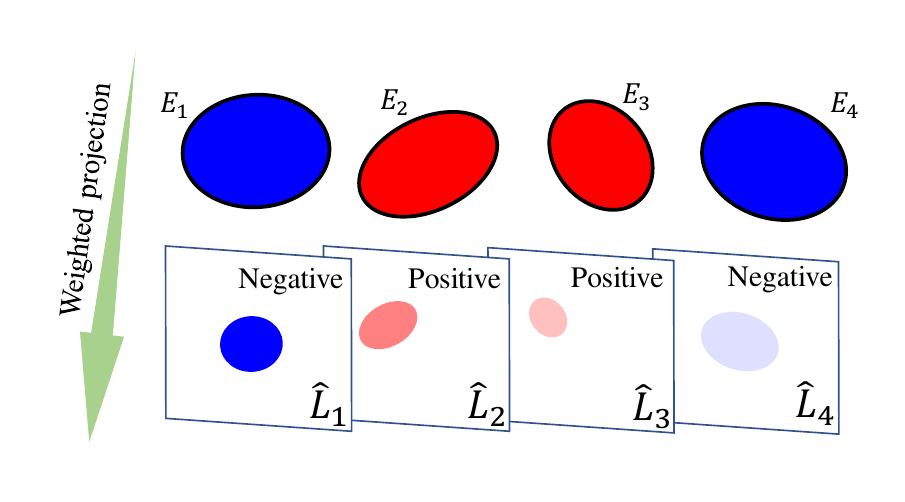}
    \label{Viewpoint_quality_evaluation_function_c}
    }
    \subfloat[]{\includegraphics[width=1.5in]{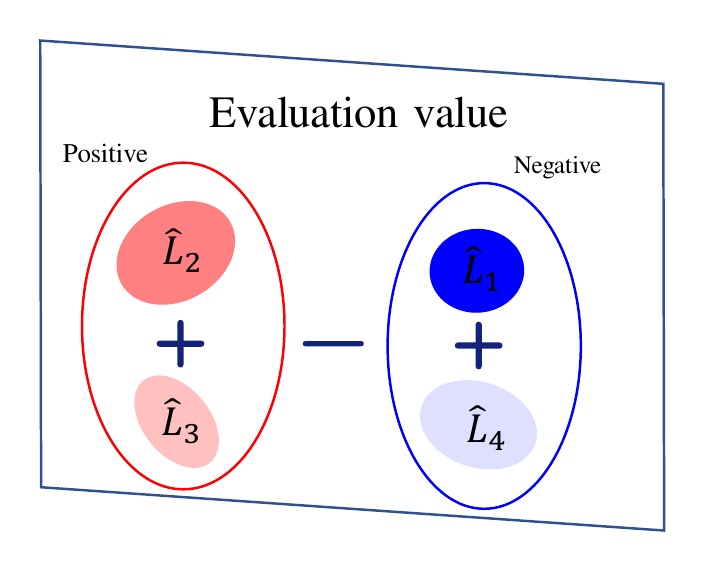}
    \label{Viewpoint_quality_evaluation_function_d}
    }
    \caption{Quality evaluation for a single viewpoint. (a) Transform each ellipsoid's center to the current viewpoint's coordinate system. (b) Sort ellipsoids by depth and assign observability weights. (c) Project each ellipsoid based on its observability weight. (d) Combine the occupied and frontier ellipsoids to obtain the viewpoint quality evaluation.}
    \label{Viewpoint_quality_evaluation_function}
    \vspace*{-1\baselineskip} 
\end{figure*}

\begin{figure}[t]
    \centering
    \includegraphics[width=3.4in]{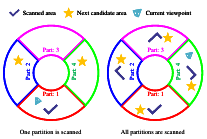}
    \caption{This figure illustrates the process of selecting the next best view based on the scanning status of different regions.}
    \label{Global_partitioning_strategy}
    \vspace*{-1\baselineskip} 
\end{figure}

\subsubsection{Ellipsoid observation weight calculation}
As shown in Fig. \ref{Viewpoint_quality_evaluation_function}\subref{Viewpoint_quality_evaluation_function_a}, using $E_o$ and $E_f$, we can determine the position $C_w$ of each ellipsoid center in the world coordinate system. With the camera's pose $H$ in the world coordinate system, we can calculate the position $C$ of these ellipsoid centers in the camera coordinate system.

As shown in Fig. \ref{Viewpoint_quality_evaluation_function}\subref{Viewpoint_quality_evaluation_function_b}, we sort all ellipsoid centers $C$ in ascending order according to their z-coordinates.
Each ellipsoid is assigned a unique depth rank $r$.
After obtaining the ranks, we calculate the observable weight $W$ for each ellipsoid as follows:
\begin{equation}
    W = 0.5^{r}
\end{equation}
\subsubsection{Ellipsoid weighted projection calculation}
Any ellipsoid $E$ in space can be described by a 4th-order square matrix $Q$, that is:
\begin{equation}
    X^TQX=0
\end{equation}
Where $X=[x \ y \ z \ 1]^{T}$.
For any given ellipsoid, the ellipse it projects onto under the action of the camera projection matrix $P$ can be represented by the matrix description $\Phi$.
\begin{equation}
    \Phi^* = PQ^*P^T
    \label{projection}
\end{equation}

For invertible quadric matrices $Q$ and $\Phi$, their dual representations satisfy $Q^{*} = Q^{-1}$ and $\Phi^{*} = \Phi^{-1}$, where the dual space encodes tangent plane (line) constraints. 
The camera matrix $P$ can be calculated by the internal parameter $K$ of the current 3D camera, and the camera's pose $H$ in the world coordinate system\cite{hartley2003multiple}.

As shown in Fig. \ref{Viewpoint_quality_evaluation_function}\subref{Viewpoint_quality_evaluation_function_c}, we project each ellipsoid individually onto the 3D camera's imaging plane, according to the eq. \ref{projection}.
Then, we calculate the sum of pixel values $L$ within each elliptical region in the image and multiply it by the observability weight $w$ to obtain the observability of the ellipsoid, denoted as $\hat{L}$.
\begin{equation}
    \hat{L} = Lw 
\end{equation}

\subsubsection{Viewpoint quality evaluation function}
To maximize the observation of frontier information from the viewpoint,
We need to prioritize the projection of larger frontier ellipsoids over occupied ones. 
For this purpose, we designed a simple evaluation function to achieve this goal:
\begin{equation}
    F = \sum_{E_f} \hat{L} - \sum_{E_o}\hat{L}
\end{equation}

As shown in Fig. \ref{Viewpoint_quality_evaluation_function}\subref{Viewpoint_quality_evaluation_function_d}, the projection of the frontier ellipsoid is the positive part of the observation quality assessment. In contrast, the projection of the occupied ellipsoid is the negative part.
Therefore, the sum of the weighted projections of all $E_f$ frontier ellipsoids minus the sum of the weighted projections of all $E_o$ occupied ellipsoids can represent the quality $F$ of the viewpoint.

\begin{figure}[t]
    \centering
    \subfloat[Simulation environment]{\includegraphics[width=1.5in]{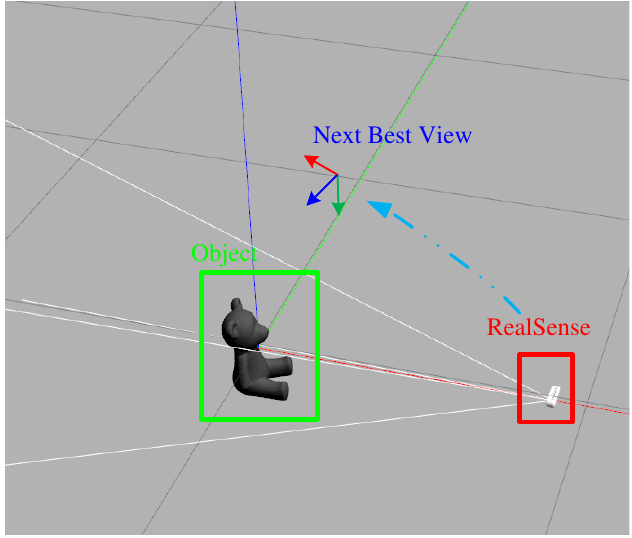}
    \label{Simulation_environment}
    }%
    \hfil
    \subfloat[55 objects to be tested]{\includegraphics[width=1.5in]{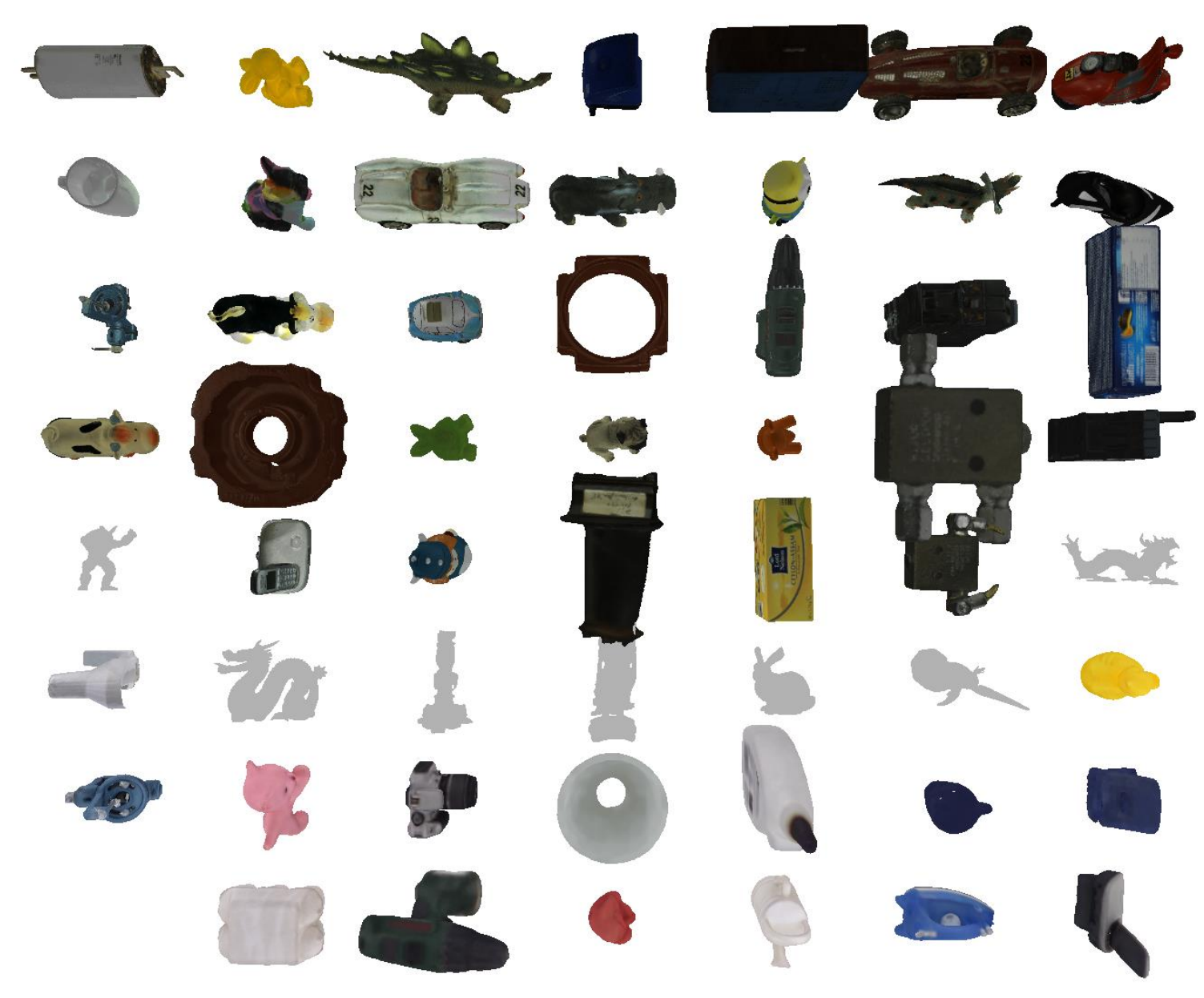}
    \label{Simulation_target}
    }%
    \caption{Simulation environments in Gazebo. }
    \label{Simulation_environment_target}
    \vspace*{-1\baselineskip} 
\end{figure}

\subsection{Global Partitioning Strategy}
Our framework needs to use the ICP\cite{besl1992method} to correct the pose of the point cloud.
Selecting the viewpoint with the highest observation quality F greedily as the best viewpoint can lead to discontinuous scanning and failure in point cloud registration. 
Moreover, an overly greedy selection strategy may lead to backtracking, resulting in a reduction in the efficiency of our framework.

To address this, inspired by the concept of region clustering\cite{lee2022uncertainty}, we introduce a global partitioning strategy.
This strategy encompasses not only the partitioning of candidate regions but also the assurance that the viewpoint can be selected among consecutive partitions, which plays a significant role in ensuring the success rate of point cloud registration. Consecutive partitions are crucial for enhancing the success rate of point cloud registration.

We divide the candidate view sampling area into $\beta$ regions according to longitude.
The selection of $\beta$ depends on the current registration strategy. 
After all regions have been scanned, the next best view can be selected from all partitions with the largest observation quality $F$.
Otherwise, the next best view can only be selected in the unscanned neighborhood of the scanned partition, as shown in Fig. \ref{Global_partitioning_strategy}.

\begin{figure*}[!t]
    \centering
    \subfloat[Coverage for each framework]{\includegraphics[width=3.4in]{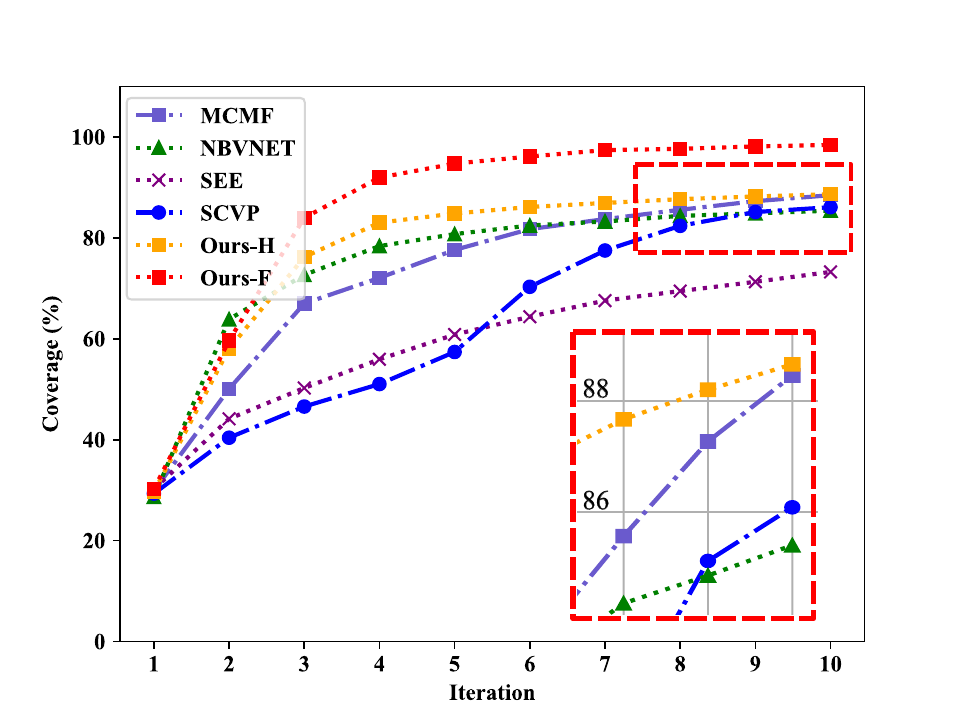}
    \label{Simulation_Comparison_Coverage}
    }%
    \subfloat[Compute time for each framework]{\includegraphics[width=3.4in]{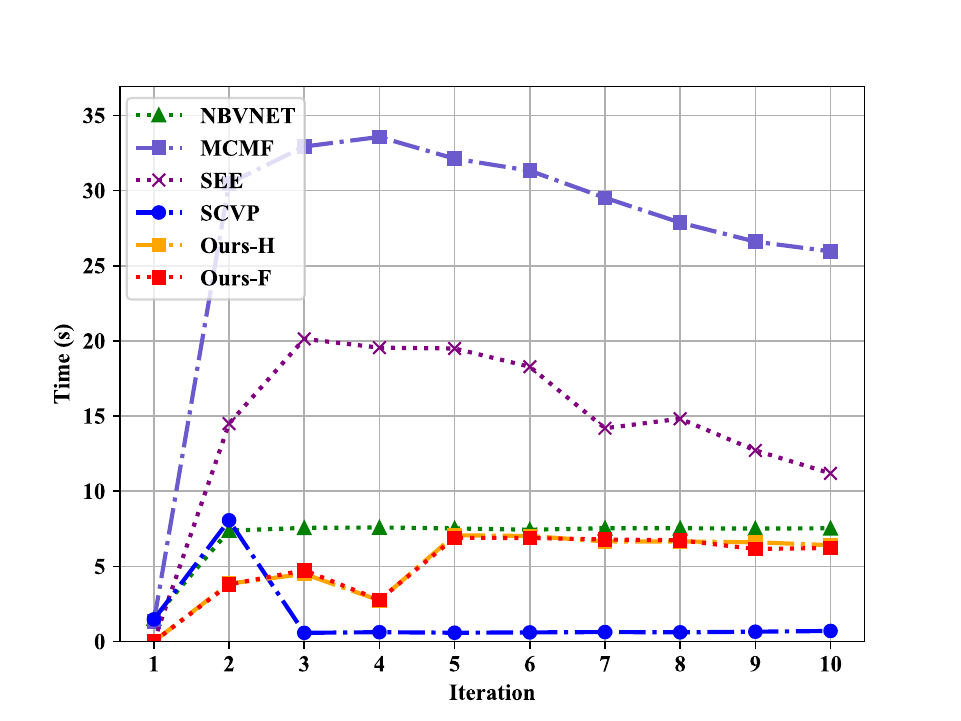}
    \label{Simulation_Comparison_Times}
    }%
    \caption{Comparison experiment tests 55 objects in the simulation environment, with results showing the average performance of all objects per iteration.}
    \label{Comparison_Result_Chart}
    \vspace*{-1\baselineskip} 
\end{figure*}

\section{Experiments}
\label{sec:result}

\subsection{Simulation Experiments}
To evaluate the performance of our framework, we constructed a simulation environment using the Gazebo platform on the Ubuntu operating system.
Computations are conducted on a CPU-only device with an i7-14700F processor.
The simulation environment is configured with a single RealSense camera and includes 55 objects selected from the Stanford 3D Scanning Repository\footnote{\href{https://graphics.stanford.edu/data/3Dscanrep/}{Stanford: https://graphics.stanford.edu/data/3Dscanrep/}}, Linemod, and HomebrewedDB\footnote{\href{https://bop.felk.cvut.cz/datasets/}{Linemod and HomebrewedDB: https://bop.felk.cvut.cz/datasets/}}.
The simulation environment is depicted in Fig. \ref{Simulation_environment_target}.

The Simulation experiment is based on two evaluation metrics:
(1) Point Cloud Coverage: Sampling 10,000 points from the model file, traversing all model points, and calculating the proportion of model points that have a point in the input point cloud within a distance of 0.005m.
(2) Computational Time: Measuring the computation time of the NBV planner per iteration, covering scene representation update and optimal viewpoint selection.

\subsubsection{Comparison experiments}
In Comparison experiments, our framework (with full-sphere and hemisphere candidate viewpoints sampling area) will be compared with the voxel-based MCMF\cite{pan2021global} (completeness-focused) NBV-Net\cite{mendoza2020supervised} and SCVP\cite{pan2022scvp} (efficiency-focused), and the point cloud-based SEE\cite{border2024surface}.
Key experimental parameters are specified in Tab. \ref{comparison_res}, with remaining parameters set to their default values.

\begin{table}[ht]
    \caption{Performance Comparison}
    \centering
    \begin{tabular}{cccccc}
        \toprule
        \textbf{Frame} & \textbf{Cov. (\%)} & 
        \textbf{T. (s)} &
        \textbf{Cand. Views} &  
        \textbf{Area} &
        \textbf{Res. (m)} \\
        \midrule
        SEE\tablefootnote{The direction SEE's viewpoint is defined by its planner, while the rest point to the object center. For resolution, SEE uses a point cloud density radius, while others use voxel size.} & 73.24 & 14.49 & 800 & Hemi & 0.03 \\
        SCVP\tablefootnote{Both SCVP and NBV-Net utilize the pre-trained model from the SCVP open-source project. As the model requires input voxels of size 32×32×32, we uniformly set the voxel resolution to 0.006 m and employed a fixed set of 32 candidate viewpoints}     & 86.08 & 1.45 & 32  & Hemi & 0.006 \\
        NBVNET   & 85.40 & 6.91 & 32  & Hemi & 0.006 \\
        \textbf{Ours-H} & \textbf{88.66} & \textbf{5.15} & \textbf{800} & \textbf{Hemi} & \textbf{0.03} \\
        \midrule
        MCMF     & 88.46 & 27.16 & 800 & Full & 0.03 \\
        \textbf{Ours-F} & \textbf{98.41} & \textbf{5.10} & \textbf{800} & \textbf{Full} & \textbf{0.03} \\
        \bottomrule
    \end{tabular}
    \label{comparison_res}
    
    \smallskip
    \begin{flushleft}
        \textbf{Cov.}: Coverage; \textbf{T.}: Computation Time; \textbf{Cand. Views}: Candidate Views; \\
        \textbf{Res.}: Resolution; \textbf{Hemi}: Hemisphere; \textbf{Full}: Full Sphere.
    \end{flushleft}
    \vspace*{-1\baselineskip} 
\end{table}

In this experiment, our global partition number $\beta$ is 4, and $ T_{max}$ is 10, and our framework processed an average of 2123.40 voxels and an average of 11.06 ellipsoids (including occupied and frontier).
All frameworks receive data from the RealSense camera and adjust the camera's pose based on their NBV outputs for iteration, starting with the same initial pose. Each framework performs 10 iterations\footnote{If any framework produces fewer than 10 views for an object, we duplicate the last view until we reach 10 views.}, with point cloud coverage and computational time analyzed as key metrics. 
This experiment uses 10 iterations as the benchmark, emphasizing the comparison of how efficiently different frameworks rapidly improve object scan coverage, rather than their ultimate convergence performance.

As illustrated in Fig. \ref{Comparison_Result_Chart}, comparative results show that our framework not only achieves the highest coverage but also maintains lower compute time compared to other benchmark frameworks. Table \ref{comparison_res} presents the comparisons in different sampling regions, revealing that our framework consistently exhibits superior coverage and better computational efficiency in both hemispherical and full spherical sampling spaces. 
The experimental results demonstrate that the ellipsoid representation structure can efficiently and flexibly characterize the spatial distribution features of frontier regions. Integrated with a projection-based optimal viewpoint selection strategy, our framework enables rapid and accurate determination of the next best observation viewpoint, thereby significantly improving the scanning efficiency for unknown objects.

\subsubsection{Ablation experiment}
The maximum number of ellipsoids \(T_{max}\), the number of partitions \(\beta\), and the initial viewpoints are all manually set. To validate the rationality of these settings, we designed three ablation experiments to evaluate different configurations.

\begin{figure*}[!ht]
    \centering
    \subfloat[Effect of $T_{max}$ on coverage]{\includegraphics[width=2.3in]{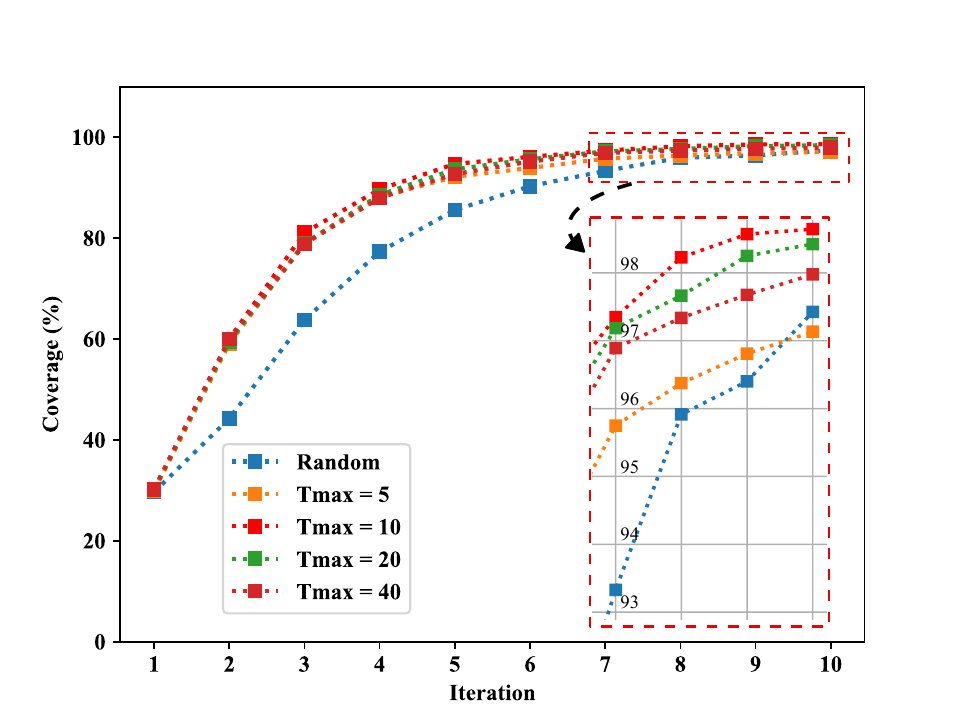}
    \label{Albation_T_max_Coverage}
    }%
    \subfloat[Effect of $\beta$ on coverage]{\includegraphics[width=2.3in]{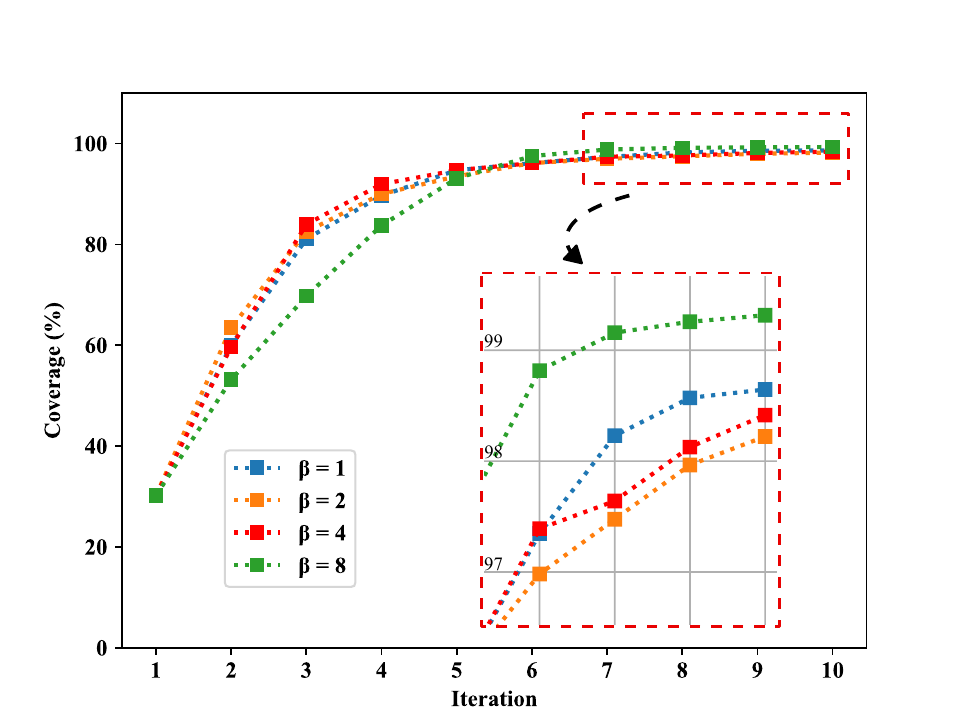}
    \label{Albation_Beta_Coverage}
    }%
    \subfloat[Effect of initial pose on coverage]{\includegraphics[width=2.3in]{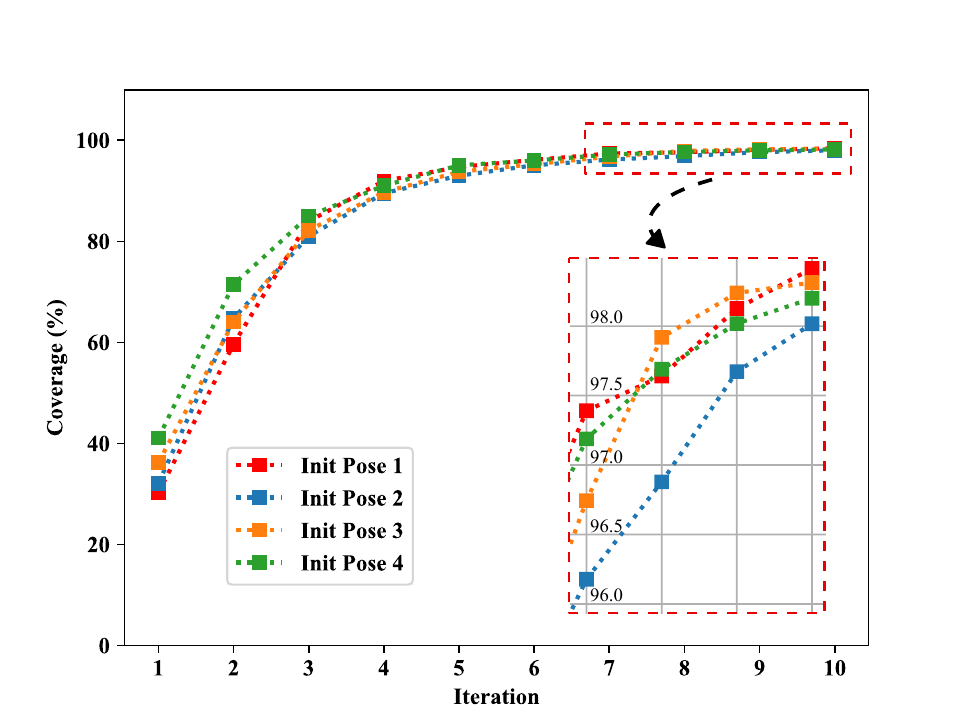}
    \label{Albation_Init_Pose_Coverage}
    }%
    \\
    \vspace*{-0.9\baselineskip}
    \centering
    \subfloat[Effect of $T_{max}$ on compute time]{\includegraphics[width=2.3in]{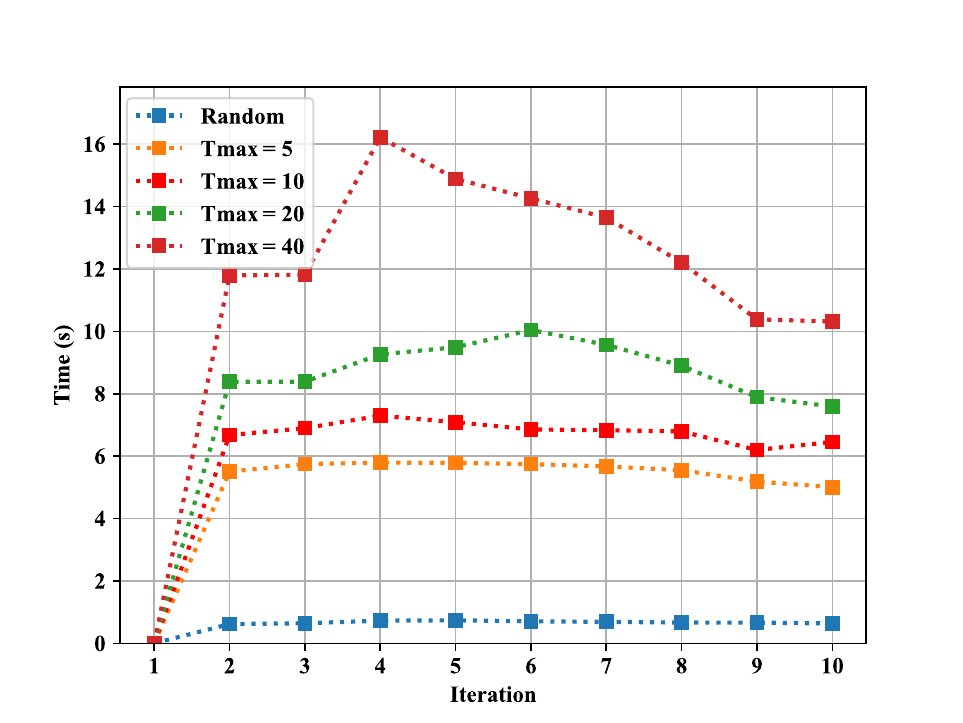}
    \label{Albation_T_max_Time}
    }%
    \subfloat[Effect of $\beta$ on compute time]{\includegraphics[width=2.3in]{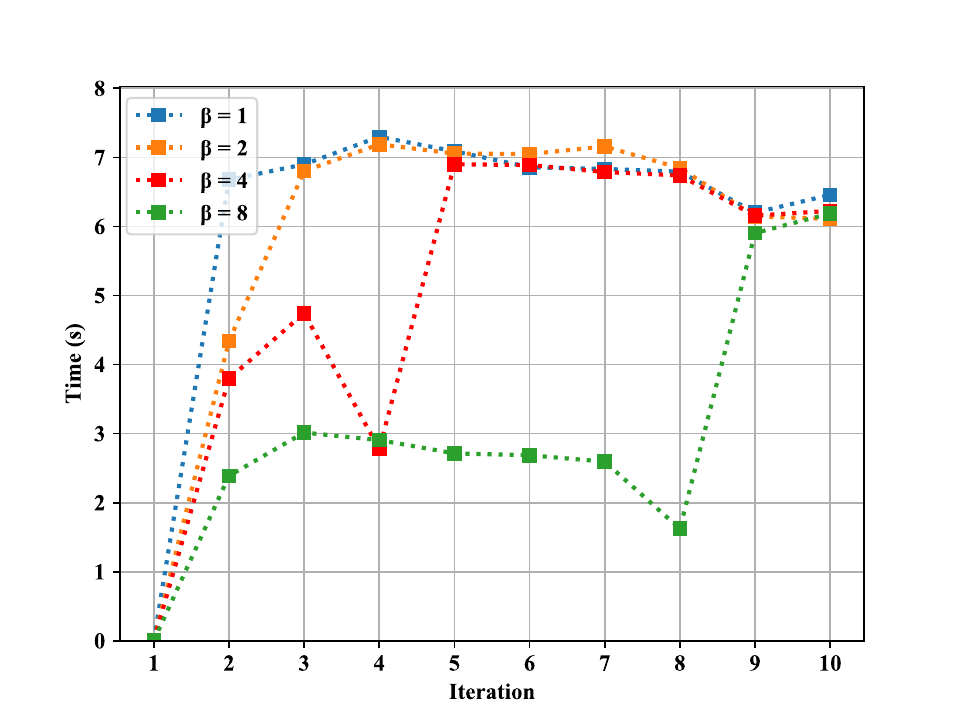}
    \label{Albation_Beta_Time}
    }%
    \subfloat[Effect of initial pose on compute time]{\includegraphics[width=2.3in]{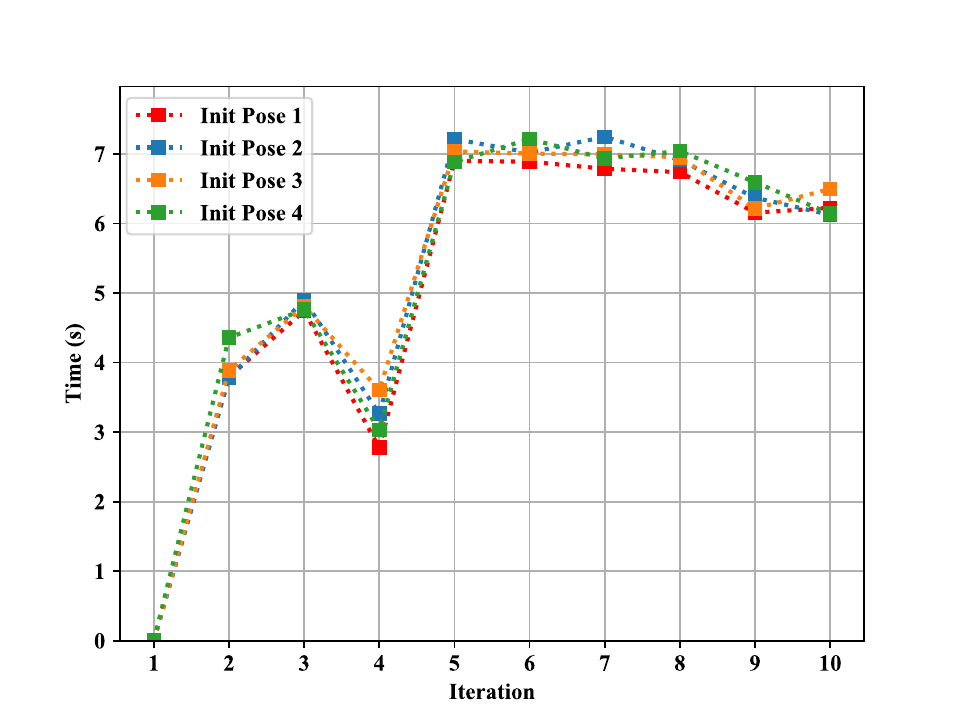}
    \label{Albation_Init_Pose_Time}
    }%
    \caption{Ablation experiment tests 55 objects in the simulation environment, with results showing the average performance of all objects per iteration.}
    \label{Albation}
    \vspace*{-1\baselineskip} 
\end{figure*}

We set \(\beta\) to 1 (disabling the global partition strategy) and varied \(T_{max}\) to 5, 10, 20, and 40, comparing them with the random selection strategy. The results shown in Fig.\ref{Albation}\subref{Albation_T_max_Coverage}\subref{Albation_T_max_Time} indicate that the strategy is not sensitive to parameter selection, as different \(T_{max}\) values have minimal impact on convergence efficiency, and increasing \(T_{max}\) leads to a higher computational burden.
When \(T_{max}\) is set to 10, we performed a comparison for global partition numbers \(\beta\) of 1, 2, 4, and 8. The results in Fig. \ref{Albation}\subref{Albation_Beta_Coverage}\subref{Albation_Beta_Time}  show that a higher partition number improves the final convergence efficiency but at the cost of a slower convergence rate.
To verify the stability of our framework under different initial conditions, we tested it with \(T_{max}=10\) and \(\beta=4\) using four different initial viewpoints: [0.9,0,0], [0,0.9,0], [0,0,0.9], and [0,0.63,0.63]. The results, as shown in Fig.\ref{Albation}\subref{Albation_Init_Pose_Coverage}\subref{Albation_Init_Pose_Time}, indicate that our framework is stable across different initial values.

\subsection{Real-world Experiments}
\begin{figure}[t]
    \centering
    \includegraphics[width=3.4in]{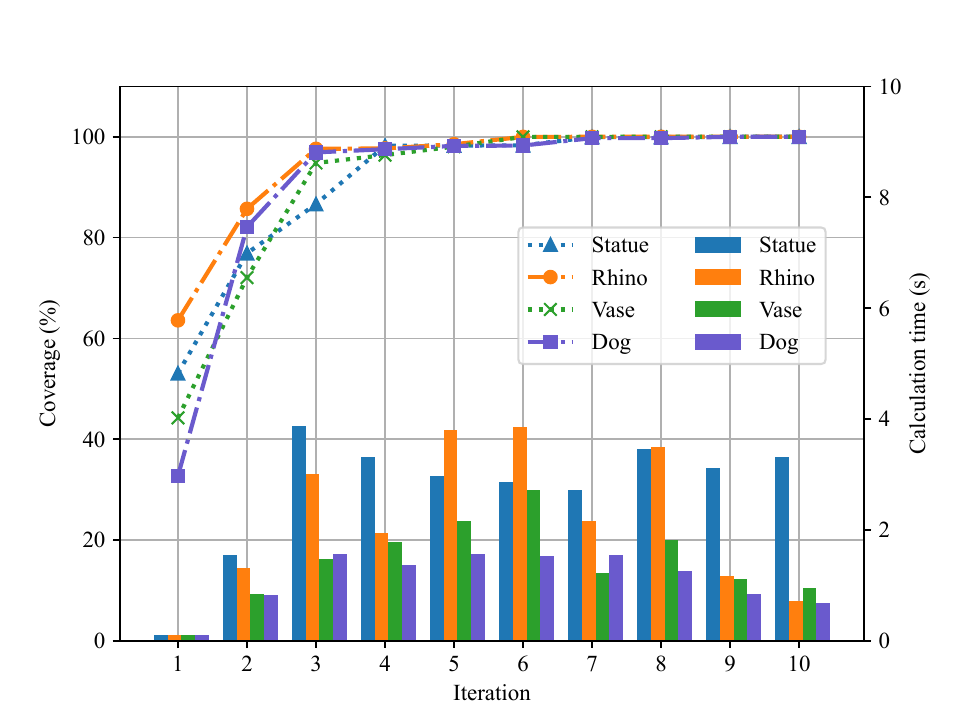}
    \caption{Chart of result data from real-world experiments. The bar chart represents the computational time for each iteration, and the line chart represents the point cloud coverage after each iteration.}
    \label{Real-world_chart}
    \vspace*{-1\baselineskip} 
\end{figure}

\begin{figure}[t]
    \centering
    \includegraphics[width=3.4in]{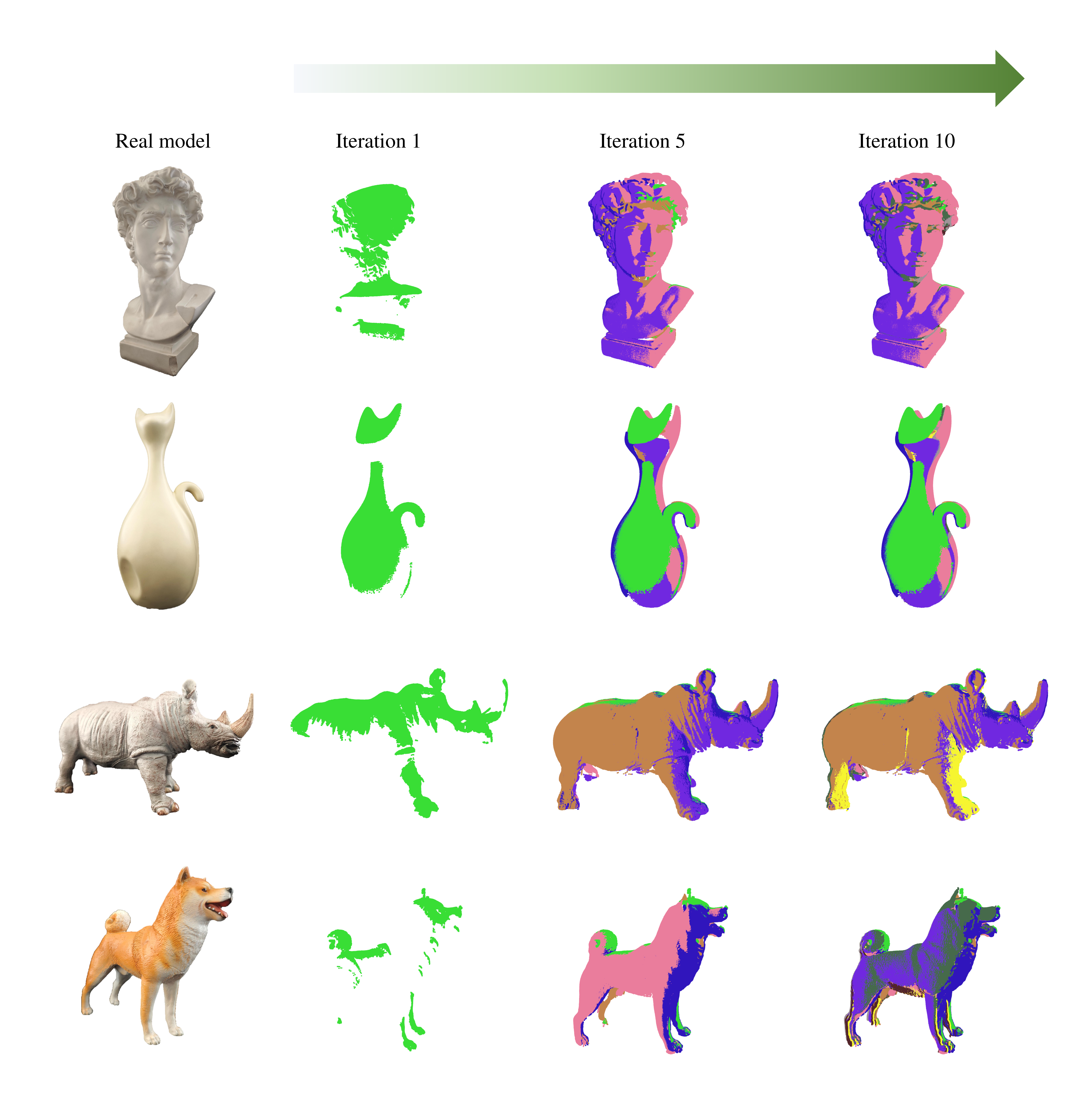}
    \caption{The models used in the real-world experiment and their reconstruction results, from top to bottom: statue (9,655 points), vase (3,013 points), rhino (3,857 points), and dog (2,684 points).}
    \label{Real-world_result}
    \vspace*{-1\baselineskip} 
\end{figure}

As shown in Fig. \ref{Experimental_platform}, our experimental platform includes a self-made structured light 3D camera, a six-axis robotic arm with 800 mm working envelope, and a 360-degree rotatable turntable to cover the entire hemispherical space of the object. 
The NBV planning framework and control programs run on a laptop with an i9-14700HX processor. 
The experiment used an Octomap resolution of 0.01m, with the number of partitions $\beta$ set to 4, the maximum number of fitted ellipsoids $T_{max}$ set to 50, the initial observation pose as shown in Fig. \ref{Experimental_platform}, and the number of iterations is 10.
Reconstruction experiments on four objects of different sizes showed that our framework could achieve good results when reconstructing different objects, proving its feasibility.
Fig. \ref{Real-world_result} presents the real-world experimental results, where the number of registered point clouds after voxel filtering (0.005m) is used to represent the object size.

Since there is no ground truth for objects in real-world experiments, we adopted the method from SEE\cite{border2024surface} by treating the final point cloud from the iterations as the ground truth to assess the convergence efficiency of the NBV process.
The iteration process is shown in Fig. \ref{Real-world_chart}.

Fig. \ref{Real-world_chart} shows that objects like the vase and dog have lower computation times. Complex objects like the statue and rhino need more ellipsoids, increasing computation times.
Although the point cloud size of the statue is more than twice that of the rhino, the computation time is similar, indicating that the computational cost of the framework is not significantly related to the object volume size.

\section{Conclusion}

This paper introduces a projection-based NBV planning framework that uses ellipsoids to represent objects in voxel structures, replacing ray-casting with ellipsoid projection to select the optimal viewpoint. This approach not only provides higher coverage but also significantly reduces computational load, with both simulation and real-world experiments confirming its efficiency and feasibility.
However, challenges remain in capturing complete surface data with structured light cameras on high-reflectivity materials or mutual reflections, which will guide future improvements to the NBV planning framework. 
In addition, efficiency-focused NBV planning frameworks should also focus on system movement efficiency, aiming for minimal movements and optimal global path planning.

\addtolength{\textheight}{0.cm}   
\bibliographystyle{IEEEtran}
\bibliography{IEEEabrv,jzzpb}

\end{document}